# Open architecture for multilingual parallel texts

*M.T. Carrasco Benitez*
*Luxembourg, 28 August 2008, version 1*

# 1. Abstract

*Multilingual parallel texts* (abbreviated to *parallel texts*) are *linguistic versions* of the same content ("translations"); e.g., the Maastricht Treaty in English and Spanish are parallel texts. This document is about creating an **open architecture** for the **whole** *Authoring, Translation and Publishing Chain* (ATP-chain) for the processing of parallel texts.

# 2. Summary

## 2.1. Next steps

The **next steps** should be:

- **Administrative organisation**: create a coordinating organisation and approach the existing organisations that might cooperate; e.g., IETF, LISA, OASIS, W3C.

- **Public discussion:** with an emailing list (or similar) to reach a *rough consensus*, in particular on aspects such as the **required specifications**. Organise the necessary meeting(s).

- **Tools**: implement some tools. This might be done simultaneously with the public discussion to better illustrate the approach and support the discussion.

## 2.2. Best approaches

To obtain the best **quality**, **speed** and the lowest possible **cost** (QSC) in the production of parallel texts, one should aim for:

- **Generating** all the linguistic versions ready for publication, from linguistic resources. Probably one of the **best** approaches.

- **Seamless ATP-chain** implementations.

- **Authoring:** *Computer-aided authoring* (CAA) tools with a controlled authoring environment; it should deliver *source texts* prepared for translation.

- **Translation:** *Computer-aided translation* (CAT) tools to allow translators to focus only in translating and unburden translators from auxiliary tasks such as formatting. These tools should have functionalities such as side-by-side editor and the re-use of previous translations.

- **Publishing**: *Computer-aided publishing* (CAP) tools to minimise human intervention.

The **open architecture** (i.e., based on **open standards)** must allow the proper **interoperability** of programs from **different** software producers. It must be possible to implement *client tools* that give the impression to the users (e.g., authors, translators) of interfacing with **one seamless** system. These client applications might be interacting (through open standards) with many *application servers* that might be from different software producers; hence the complexity can be hidden from the users.

This open architecture must be designed for using Internet technologies. It should be viewed like an extension of the intranet/Internet to achieve wider interoperability with unrelated systems such as an internal financial management system or external *record management* system ("archiving"); office automation systems are mostly based on intranet/Internet technologies.

# 3. Rationale

## 3.1. Approach
This document combines:

- **A primer** for open architecture, parallel texts and related fields.
- **A proposal** for an open architecture for parallel texts encompassing the whole ATP-chain.
- **An approach** to attain these objectives.

The document is addressed to a **mixed** audience: parallel texts are at a crossroad of several fields (no pun intended). Some concepts might be self-evident to people familiar with one field and new to people in another field; hence, the style of **re-stating the evident**. The intention is to allow as a wide audience as possible to participate in a consensus process, with common terminology.

To start with, some concepts are simplified and unexplained, though they should be sufficiently clear by the context: they are explained later in the document and the gory details in the references. This is the principle of *"lie, if it helps"*:

> *"Another noteworthy characteristic of this manual is that it doesn't always tell the truth."*
>   Don Knuth, *The TEX book*.

## 3.2. Quality, speed, cost (QSC)
These are three dimensions used to measure the production of parallel texts: one should aim to have the **best** QSC; i.e., high quality, high speed and as low a cost as possible.

## 3.3. Objective
The objective, mission statement and scope is:

> *"The production of parallel texts with the **best QSC**"*

It is crucial to consider the **whole** process in order to have the best QSC in the production of parallel texts; and not only to optimize one phase, such as translation. One must be able to construct **multi-vendor seamless systems** that could **evolve for an indefinite period**. The components from the different software producers must be able to achieve **interoperability**.

## 3.4. Strategy
The strategy is to use **open standards**. In a parable: a standard is a *recipe* and a program is a *meal*. A cook might have a secret recipe that he only is willing to prepare and serve in his restaurant: his interest lays in maximising his profit by using his monopoly on this exquisite meal; the interest of the eater lays in having an **open** recipe, so he can prepare it or ask any number of cooks to prepare it.

The **programs** ("meal") could be proprietary or open source; but **standards** ("recipe") such as formats and protocols, must be **open**. A strategy based on closed standards creates many inconveniences for the clients. For example, it makes clients **captive** (*"you can only have this meal* *in my restaurant*"); hence it is very painful and costly to change to a new product as all users, potentially thousands of them, would have to be trained for the new product.

## 3.5. Private specification and open standard

One might create a private specification for a private system. But one should look around to check if the private needs are really private or generic. With little or no additional effort and cost, it is possible to write a private specification as a proper (open or closed) standard.

Designing a private specification as a proper standard would make the resulting work more solid, independently if the result is published or not, as one has to look at the private needs from a more formal perspective. One has to use and interoperate with existing standards; indeed, one has to avoid re-inventing the wheel and one has to produce a new standard only if there is a genuine gap in standards.

Through this exercise, one might discover off-the-shelf programs that fulfil the private needs with parametrisation or minor modifications. Otherwise, it might be implemented in-house or outsourced; in both cases the task should be easier by having a proper standard. If the market is sufficiently horizontal and the standard open, there are good chances that software producers would implement several competing systems according to the new open standard.

There are some cases when one wants to keep the private needs private. For example, when the advantages of a firm depend on owning private software. But in general, particularly in the case of non-software firms, one is better off by collaborating in the creation of open standards. Software firms might try to create a monopoly, but today general awareness makes it difficult (not impossible) to corner a new market sector with monopoly practices: it is a different matter with already established monopolies.

## 3.6. Present

In this field, present products and initiatives concentrate on one of the phases (e.g., translation), perhaps adding as a second thought some aspects of the others (e.g., authoring). But the proper thing to do is to consider the three phases together.

For example, for computer-aided translation (CAT), there are no comprehensive standards. Programs from different vendors are incompatible; this means that clients are forced to buy a suite from **one** vendor. Initiatives such as *Translation Memory eXchange* [TMX] are a step in the right direction, though insufficient as they only cover one out of many aspects; i.e., one needs specifications to cover the **whole ATP-chain**.

# 4. Scenarios and examples

## 4.1. Generation of parallel texts

This is creating parallel texts programmatically ready for all the required languages and formats; e.g., a text in all the required languages *camera ready* and *web ready*. Generating parallel texts should be considered at least when there are *tight phase binding* and *high reusability*. In general, humans are better at translating and machines at auxiliary tasks. This is one of the **best methods** to create parallel texts with a **very good QSC** (quality, speed, cost).

With this approach, it could be considered that the three phases of the ATP-chain are nearly woven into one. The main conceptual components for the generation of parallel texts are:

- **Linguistic table**: a data structure where each *linguistic records* contains one parallel texts (each field contains a *linguistic segment*), plus at least a *record identifier*.

- **Document template**: specification with record identifiers.

- **Program**: it generates all the linguistic versions by replacing the record identifiers in the document template by the corresponding linguistic segment from the linguistic table.

This is a simple idealised model; in practice, there would be many linguistic tables, templates and programs. The table would usually be filled with human translated segments that could be used many times over. A demonstration is the *Generation of Multilingual Parallel Texts with XML* [PARTEXT].

The conceptual approach is similar to the *Darwin Information Typing Architecture* [DITA]: *topics* ("units of information") are combined by *maps* ("documents"). A topic is a very large *segment* and hence it has a very high *semantic safety*. Indeed, one could visualize a DITA-like approach to generating parallel texts. In this document, *topic* refers to the general conceptual approach; DITA or others.

## 4.2. Continuous publication

It is a publication with **minimal** changes in comparison to the previous version. One should apply *generation of parallel texts*. A *multilingual content management system* (MCMS) is a good example; it might be designed as a very integrated ATP-chain system.

Web sites are the shop windows of organisations and visitors expect them to be up to date. The construction and maintenance of multilingual web sites is particularly tricky. It is quite different from the more traditional parallel texts. For example, blatant errors must be **corrected immediately**. If an error appears in four pages and the site is in twenty-three languages (the number of official languages in the European Union [ISG]), it means changing ninety-two pages. In addition, most translation departments are geared to the translation of documents and not for short segments; hence, the whole business soon becomes a nightmare.

## 4.3. Periodic publication

It is a publication with a **significant** content from the previous version. One should apply *generation of parallel texts*. An example of this type of publication is the European Union Budget [EUB, SEIBUD, TEC]. A *generic generator for parallel texts* adaptable to the generation of different publications (as opposed to an application for one single publication) should be developed. Also, one can apply *periodic publications pre-processing*.

## 4.4. Seamless Legislative System

In a simplistic way, for countries (and political entities) with several languages, the *final products* of the legislative process are multilingual laws; i.e., parallel texts. Different parts are under the control of different authorities. Though such a system would continue to be composed of independent sub-systems, users in each authority should have the feeling of using **one seamless** system.

There should be an **overall malleable architecture**: the requirements and technology change within months. The sub-systems must follow common **specifications** to achieve **interoperability**. Each phase should take a few months and it should deliver **concrete benefits**. One should avoid a big bang of an over-specified monolithic system: "*Trust me, I will come back in a couple of years with your wonderful system*". Reality is often different: five years late, over-budget and buggy.

Let's assume the following scenario with these imaginary programs that could come from different software producers; they achieve **interoperability** by using a common format: a *Multilingual Electronic Dossier* [MED].

### 4.4.1 Computer-aided law drafting (CALD)

This is an **integrated** *authoring system* that assists with the drafting of new laws, with facilities such as:

- **Controlled templates** for creating the new laws.

- **Connections to multilingual normative memories** [CA] to find appropriate (mandatory, recommended or verified) segments that have already been translated into other languages. Hence, no translation would be required for these segments. This could be particularly useful to non-native speakers of the drafting language.

- **Topic** to maximize reusability.

- **Authorship checker** that can give overall feedback to improve the text and facilitate translation and publication. Analogous to a spelling checker.

- **Background documentation module** to manage relevant pre-studies for this new law and similar background documents, which can be included in the MED or just linked to an external server.

- **Administrative module** to manage metadata, workflow (e.g., translation request) and packaging all the components (source document, background documents, metadata, etc) into one MED to forward for translation.

### 4.4.2 Translation pre-processing

Upon arrival for translation, *translation memories data* and additional background documents specific to the newly drafted law will be added automatically to the MED. These come from *linguistic databases* such as Euramis.

*Periodic publications pre-processing* is more sophisticated: one has to process the present version of the source language, the previous versions of the source and translated languages, and the linguistic database.

### 4.4.3 Computer-aided translation (CAT)

This is an **integrated** application for translators. It has an inbuilt *parallel text editor* that shows the source and a target text side by side; for easy identification, each segment is numbered and it has visual clues. This application could connect to servers to verify the translation quality on demand, "*look over the shoulder*" to other ongoing translations, connect servers such as Wikipedia, etc; e.g., the Spanish translator could look at the ongoing work of the French translator, both translating from English. This should be one of the first implemented tools (see *First Tools*).

### 4.4.4 Computer-aided publishing (CAP)

It is a program that produces automatically the publications in all the languages and formats required. Producing *HyperText Markup Language* [HTML] can be done automatically. Producing high quality typography can be done **almost** automatically, but it usually requires human intervention; e.g., human indicated refinements following *badness* feedback in TeX [TEX].

## 4.5. Textual interpretation

*Textual interpretation* is *transforming* (same language or translation; in real-time or afterward) any *natural language form* (e.g., speech, sign-language, text, etc) to text. For example, transforming a Spanish speech into an English text projected into a large screen in a room in real-time; there are two transformations: the *interpretation* ("translation") from Spanish to English and the *form* from speech to text.

The main types of textual interpretations are:

- **Scribe**: in real-time, transforming speech to text in the same language. Even without interpretation, this greatly improves the understanding of non-native speakers; e.g., scribe in some ICANN meetings [ICANN].

- **Interpretation**: in real-time, transforming speech from a language to text in different language(s). Similar to traditional interpretation.

- **Consecutive**: in near real-time, transforming text from a language to text in different language(s). Similar to consecutive interpretation.

- **Translation**: afterward, transforming text from a language to text in different language(s). Similar to traditional translation.

For each type of textual interpretation, the most common natural language form input is indicated, although the input could be in any natural language form. Indeed, there are also variations such as:

- **Pivot form**: using an intermediary natural language form; e.g., for interpretation, first scribe (to text) and immediately interpret (to text) to the target language(s).

- **Pivot language**: using an intermediary language; e.g., from French speech to English text and from English text to Spanish text, so the pivot language in this case is English.

- **Interaction**: for example, a Spanish reader of an English text could request the simultaneous textual interpretation of one English sentence.

- **Summary**: as in translation, there could be a summary of the source language and a translation of this summary into several languages.

Textual interpretation depends on *chat* techniques such as *Extensible Messaging and Presence Protocol* [XMPP]; this protocol has a draft extension for *Language Translation* [XEP-LT], though it probably needs more work.

One might have a *textual interpretation hub* (i.e., far from the meeting room) that provides a service to many meetings. For many meetings this approach is sufficient. A *textual interpreter* is a new type of professional that requires a new mixture of skills.

# 5. Policy

The main considerations are *policy factors* and not *technical factors*. Even minor policy decisions can have far greater impact than big technical advances; e.g., the simple decision of not translating a whole category of documents is more significant than a wonderful new electronic dictionary. Policy addresses aspects such as:

- **Accounting**: measuring the **total** cost and **partial** costs, to apportion the costs properly.
- **Constraints**: what to enforce in each phase to have a better whole.
- **Entirety**: what is produced; e.g., full translation or only a summary.
- **QSC (quality, cost, speed)**: e.g., human or machine translation; printout or proper typography.

## 5.1. Accounting
One should have proper accounting for the total and for **each phase** of the ATP-chain. To apportion the costs properly, one has to identify the **phase's boundaries**.

For example, translation is to perceive (usually reading) a text in a language and expressing it (usually writing) in another language; all the rest is either authoring or publishing. Translating a badly written source text in an awkward file format must carry additional cost to the authoring phase; i.e., the translation phase must not pay for the sloppiness of the authoring phase. Formatting (for the web or other) or typography is publishing.

## 5.2. Constraints
Constraints refer to the limitations in each phase of the ATP-chain in order to facilitate the overall process with the objective of obtaining the **best** QSC (quality, speed, cost).

One might require that documents (e.g., 500 pages) must have a summary (e.g., 5 pages) that might be used for *summary* translations; indeed, the author is the best person to make the summary. If the source document does not contain a summary, it must be produced in the source language (by a human or a machine [SB]) **before** translating. Hence: the task of summarising is done only **once**; it would be the **same** summary for all linguistic versions.

*"Translation starts with authoring"*: a source text **prepared** (written, structured, formatted, etc.) with translation in mind is far easier and more economical to translate. The main aspects are:

- **Linguistic**: the language style itself; e.g., simple grammar.
- **Structure**: the document structure, including *metadata*; one can view it as the *content model*.
- **Format**: the file format, including *content and presentation*.

Indeed, one can use a topic approach.

## 5.3. Entirety
This is the policy decision on what has to be produced (and when), particularly in the translation and publication phases. For example:

- **Suspended translation**: translate only when certain conditions are fulfilled; hopefully never.

- **Summary translation**: translate only a 5 pages summary out of 500 pages.

- **Electronic publishing**: decide that the statutory final publication is electronic. Other non-statutory parties could be allowed to sell the paper version, under certain conditions.

Stating the obvious: translation has a **cost** and the lowest translation cost is achieved by **not translating**. The cost in 2005 was about 511 million euros [SR] for the European Parliament, the European Council and the European Commission; i.e., excluding all the other European institutions and bodies.

## 5.4. QSC (quality, speed, cost)

This refers to the quality and speed required in total and for each phase of the ATP-chain (authoring, translation, publishing): usually, the more **quality** and **speed** are required, the higher the **cost** (QSC). For example, human or machine translation (translation quality); the final text is required in 24 hours (total speed); printout or proper typography (publishing quality). In brief:

quality + speed = cost

# 6. IT architecture

The intention is the **interoperability** among programs from different producers and the strategy is **open standards**. The approach should be seen as **extending** the Internet technologies to cover new application domains; hence, one shall follow the architectural practices [ARCH1, ARCH2, ART] of organisations such as the *Internet Engineering Task Force* [IETF], and the *World Wide Web Consortium* [W3C].

It is based on the application layer, like email [SMTP, POP] or the web *Hypertext Transfer Protocol* [HTTP]. And around the components: *identifier* (e.g., URI), *format* (e.g., HTML), and *protocol* (e.g., HTTP).

One should specify the open **standards** (e.g., format, protocol), and not the details of a particular **program** implementation, that could be of different richness. For example, there is no specification on how a browser should look like or which functionalities it should contain: some browsers are just *command line* and others are highly graphical with a myriad of features; however, browsers must follow the relevant specifications such as HTTP. When designing IT systems, one has to be:

- **Humble**: most of IT projects end in disaster [MC, NPfIT1, NPfIT2, VC].

- **Bifocal**: look simultaneously at short-term pressing needs and long-term strategy ("*plan big, start small*").

- **Defensive**: parallel texts systems might have to interoperate with other future or legacy information systems; e.g., versioning system or long-term archival system [LTANS, DOCPRE].

The client tools should emulate the email clients and browsers: one has a wide choice and they are well separated from the servers.

# 7. First tools

These are the first tools that should be implemented. One might start in the middle phase (translation) with computer-aided translation (CAT) tools. There should by at least *two independent implementations* (an IETF practice) of CAT tools with the following characteristics:

- **Parallel texts editor**: it must be included.

- **Formats of the parallel texts**: implement the easiest of the suitable formats (e.g., HTML, RTF, plain text).

- **Format of the translation memories**: it must be TMX.

It would be desirable to have a *web-based CAT tool*. A *Linux, Apache, MySQL and PHP* [LAMP] approach or similar should be considered, though it could also be implemented as one self-contained implementation; e.g., CVStrac [TRAC] contains in **one binary file** the web server, database, wiki, repository browser, etc. Web-based has advantages such as: no binaries in the client computers; it should be easier to create a cooperating system as the data is already on a server. With binaries in the client computer, one would need a server for the common data. It should also be explored using the newer web approaches such as Prism [PRISM].

It should be possible to implement the CAT tools by modifying open source editors and wordprocessors. For example, Abiword, NVU, OpenOffice, etc [AB, NVU, OO]. The CAT tools could be partial implementations: the objective is to have as **soon** as possible tools to **illustrate** and **validate** the general approach. Though, the implementations should be modular enough to allow adding new functionalities later; e.g., a translation quality checker (inbuilt or server-based).

Hence, one should avoid all unnecessary complications: the format(s) of the parallel texts must be the easiest to implement as per the existing source code. For example, for HTML it might be NVU, and for *OpenDocument Format* [ODF] (RTF or plain text) it might be Abiword or OpenOffice.

Another useful tool easy to implement would be a *MED validator*: a checker [LINT] and generator of the main index file from the MED content.

One has to encourage **commercial** software producers to develop high quality **interoperable** tools: some might be full suites; others specialised modules.

# 8. Authoring, Translation and Publishing Chain (ATP-chain)

These are the three phases in the production of parallel texts. Traditionally, authors just concentrate on authoring, translators on translating and publishers on publishing. But the ATP-chain has to be considered as a **whole** and one has to seek the **best QSC** (quality, speed, cost) of the production of parallel texts. For example, one might consider increasing the quality in the authoring phase in order to facilitate translation and to attain a better global QSC.

The ATP-chain must be seen as **bi-directional**; i.e., translators should be able to request authors to modify the source text to facilitate translation, and the same goes for publishers (making request to authors and translators). Each of these phases contain smaller steps. For example, the creation of summaries or the identification of segments is part of authoring; *record management* ("archiving") [MoReq, LTANS, DOCPRE, RM] is part of publishing.

## 8.1. Phase binding
It is the degree to which the phases of the ATP-chain can be integrated for the production of parallel texts. Often, phases cannot be integrated for administrative reasons, though technically it could be possible. The two extremes in a continuum (i.e., some cases fall in between) are *loose* and *tight*.

## 8.2. Loose phase binding
The interoperability is complex. **Interfaces** are required to connect the phases; e.g., an *electronic dossier* (container format). This is the more general case. The traditional relation between the author, translator and publisher is loose.

## 8.3. Tight phase binding
The phases could be integrated even to form a **monolithic** system. An example is *multilingual content management system* (MCMS) for web sites.

# 9. File formats

The file format largely conditions linguistic tools.

## 9.1. Content and presentation

Ideally, *content and presentation* should be separated. Among others it facilitates: the identification of the *plain text* and hence the processing along the ATP-chain; and the generation of multiple presentations of the same text.

In addition to presentations such as content for a web site, there should be presentations for aspects such as the *automatic harvesting* of segments for inclusion in *linguistic databases* (e.g., Euramis) and for *record management*.

XML is a good example of **content** technique and *Cascading Style Sheets* [CSS] of **presentation**. The *Rich Text Format* [RTF] is a good example of a format without separation of content and presentation.

## 9.2. Classification

For the issues at hand, text formats are classified as:

- **Plain text**: it just contains natural language.

- **Formatting-based:** it contains natural language and additional *presentation* information; e.g., RTF. Tricky to parse ("chop") for content.

- **Markup-based**: it contains natural language and additional *content* (structural) information; e.g., XML. Straightforward to parse.

# 10. Data structures for natural language

## 10.1. Linguistic segment

Abbreviated to *segment*. It is a *unit of language representation*. It can be a fixed language representation (paragraph, sentence, term, etc) or meta-language representation (a grammatical construction, machine translation coding, etc). More general, a segment is a discrete linguistic unit whose meaning is created by the program processing it. Segment corresponds roughly to *translation unit variant* in TMX, (`tuv` element)*, linguistic object* in Euramis*, chunk*, etc.

Segment as defined here must be usable for translation memory ("exact strings"), machine translation ("grammatical representations") and other natural language techniques. A parable is *regular expressions* (the *grep* command in Unix): it encompasses literal strings ("exact strings") and more advanced pattern expressions.

Similarly to *Extensible Markup Language* [XML] elements, segments can contain other segments; e.g., a paragraph would probably be composed of several sentences.

## 10.2.  Data types

### 10.2.1 Monolingual data

In addition to the segment (the basic component), the two main types of monolingual data are:

- **Text**: continuous natural language text, usually in one file. One seeks to **identify** the segments within the text. An illustration is a narrative document.

- **List:** discrete segments. They do not form a continuous discourse. An illustration is a monolingual dictionary.

### 10.2.2 Multilingual parallel data
Two or more *linguistic versions* ("translations") of the same content combined to form parallel data. The main classes are:

- **Parallel texts**: two or more files, each with a different linguistic version. An illustration is two narrative documents that are translations of each other; e.g., one in English and the other in Spanish. One seeks to **align** the identified segments in each linguistic version. This is the partner of monolingual *text* above.

- **Linguistic table**: two or more synchronised lists as *linguistic records*, each in a different linguistic version. An illustration is a bilingual (or n-lingual) dictionary. This is the partner of monolingual *list* above.

### 10.2.3 Language neutral data
It does not contain natural language. For example, the same data can be included in any linguistic version: an annex where the headers are labelled with number, which are explained in another section. This is also called *no linguistic content*. As much as possible of the data should be prepared as language neutral as it is a very effective and **economical** approach.

## 10.3. Multilingual file
According to the moment of processing, there are the following main types:

- **During preparation**: it makes life difficult.

- **Final publication**: e.g., for example printing side-by-side two monolingual texts. One can view it as a form of presentation. Not problematic.

During the preparation, there are the following types:

- **A very mixed** multilingual file has a significant amount of text in several languages; e.g., the *panache* documents of the European Parliament.

- **A serial** multilingual file is usually bilingual and it has the main text in one language (e.g., Spanish) and one or several sections (e.g., annex with lots of numbers) in a second language (e.g., English) with limited amount of natural text. The cost of translating is low, but the publishing cost (typography) is high.

## 10.4. Monolingual text

### 10.4.1 Indicator versus markup
The characteristics are:

- **Indicators** are characters **part** of the natural language text. They are used to **guess** the segment boundaries in the text; e.g., *dot* is used to guess the end of a sentence.

- **Markups** are characters **not part** of the natural language text. Hence, **no guessing** here; one knows for certain where the segment ends.

*Separator* is a type of simple markup that indicates just the boundaries between segments, but it does not carry structural information; i.e., it does not create a *tree* as in XML. The text is considered a flat string with separators; this text might be a file or just content of an XML element. This is particularly useful for *sentence segments* and *sub-sentence segments* and it greatly improves parsing. Separators might be just one character not used in the natural language, by definition.

### 10.4.2 Programmatic segmentation

Abbreviated to *segmentation*. It is identifying segments in a text with a computer program. The two main types of algorithms are:

- **String:** it uses *string analysis* and relies on indicators and markups. The algorithms are simpler and they are usually language **independent**; the same algorithm can be used for many languages [TB].

- **Grammatical**: it uses *grammatical analysis*. The algorithms are more complicated and they are language **dependent**: different algorithms must be used for different languages.

Most algorithms are mono-text; i.e., they do not take advantage from parallel texts. Segmentation algorithms using parallel texts should be better, particularly for string analysis. But even grammatical analysis should benefit, particularly among closely related languages.

### 10.4.3 Manual segmentation

It is identifying segments in a text by a human. This is rare, except when the text has to be processed by human **for other purposes** and identifying segments cost practically nothing; indeed, identifying segments might help the human with his main task. Manual segmentation is realistic when well-adapted tools are available; e.g., the tools suggest the segmentation and the human can easily accept or modify the suggestion.

The trivial example is manual segmentation at the moment of authoring; indeed, as with the summaries, the best person to segment is the author. Other examples are at the moment of translation, translation revision or proof reading for publication. Sometimes, this high quality manual segmentation is unintentionally produced and discarded; e.g., a translator might have to identify segments to use a tool and once the translation is completed this valuable manual segmentation is cleaned out and not kept for later processing such as inclusion in a linguistic database.

The term *manual segmentation* should not be abbreviated: *segmentation* on its own refers to *programmatic segmentation*.

### 10.4.4 Segment size

This classification reflects a historical thinking in term of formats without markups (i.e., using *indicators*), though it also applies to other formats:

- **File segment**: a trivial case as it is the content of one file or topic.
- **Paragraph segment**: hinted by a *paragraph indicator*; e.g., a blank line.
- **Sentence segment**: hinted by a *sentence indicator*; e.g., a dot, semicolon or similar.
- **Sub-sentence segment**: it usually requires markups, though one could try grammatical analysis on its own or combined with string analysis.

### 10.4.5 Text granularity

It is a property of text that refers to the level of detail that segments in a text can be *programmatically identified*. Different parts of the text might have different text granularities; e.g., a file might have a *full sentence granularity* (i.e., the whole file can be segmented at sentence level) or *partial sentence granularity* (i.e., only some sentences in the file can be segmented at sentence level).

### 10.4.6 Relation to XML

Expressing these concepts in XML is quite simple. Indeed, processing a densely (i.e., including sub-sentence segments) marked XML document would be straightforward. But often one has to deal with documents with little or no markups.

## 10.5. Multilingual parallel texts

Abbreviated to *parallel texts*. These are *linguistic versions* of the same content; e.g., the Maastricht Treaty in English and Spanish are parallel texts. Parallel texts are a metaphor on parallel lines where each line represents one language, though some lines might be "broken"; e.g., one linguistic version might be a *partial translation*.

### 10.5.1 Segment semantics

By definition, the semantics in each linguistic version must be the same. The segment might be just a string where there is no further analysis or attempt to knowledge representation. One has to consider:

- **Own semantics**: It is having a meaning on its own.

- **Context semantics**: It is having the proper meaning in relation to the whole text. It increases when the text has an associated domain; e.g., law, computers, etc. Indeed, there could be finer attributes.

One seeks *semantic safety*; i.e., using the segment with the intended semantics. This is particularly important for the generation of parallel texts and it is related to *one to many records*. The larger the segments, the safer they are; indeed a topic should be totally safe. *Sentence* is the smallest segment that has a good chance of being semantically safe. *Sub-sentence* is a too broad classification and it is usually not sufficiently safe.

### 10.5.2 Parallel texts entirety

Abbreviated to *entirety*. It is a property of parallel texts that can indicate several attributes: completeness, quality, etc. The entireties are: *complete*, *partial*, *summary*, *translating*, *machine*, *suspended*, and *undefined*.

Several attributes can be combined. Examples:

- **complete**: the English and Spanish linguistic versions are complete translations of each other.
- **partial**: the French version excludes a large annex.
- **summary**: the German version is just a summary.
- **translating**: the Dutch version is in the process of being translated.
- **machine**: the Chinese version was done by machine translation.
- **suspended**: the Russian version could be translated after three departments have requested it, so the cost to each department would be a third.
- **undefined**: the Estonian version is in an undefined state.
- **summary + machine**: the Farsi version is a summary done by machine translation.

As already stated in the policy section: the less is produced the lower the cost. For example, the decision of producing only summaries from large documents would represent huge saving.

### 10.5.3 Parallel texts alignness
Abbreviated to *alignness*. It is a property of parallel texts that refers to synchronising the segments among parallel texts.

### 10.5.4 Parallel texts granularity
It is a property of alignness that refers to which detail segments in parallel texts can be programmatically aligned. The classification of the parallel texts granularities mirrors the classification of the segments: *file*, *paragraph*, *sentence* and *sub-sentence*.

The alignness depends on the text granularity of each linguistic version. As with *text granularity*, different parts of the parallel texts might have different *parallel texts granularities*; it cannot be finer than *text granularity* of each file; e.g., for a given part of the parallel texts, if the English version has a text granularity at *paragraph* level and the Spanish version has a text granularity at *sentence* level, the parallel texts granularity can be only at *paragraph* level.

### 10.5.5 Multilingual file alignness
This is when one of the linguistic version files is multilingual. For example, one file is multilingual (English, French, German) and the other five monolingual files (English, French, German, Spanish); the aligned English segments from the multilingual file should be the same as corresponding segments from the monolingual English file. This is very much a real case and not a farfetched example; indeed, it could be worse: when the multilingual file already contains aligned text.

## 10.6. Tabular structures

### 10.6.1 Linguistic table
It is a data structure where each *linguistic record* (row) is parallel texts, plus additional non-segment data. A minimal example:

| Number | English | Spanish |
|--------|---------|---------|
| 1 | hello world | Hola mundo |
| 2 | white cat | gato blanco |
| 3 | white cat | gata blanca |

Linguistic record is abbreviated to *record*. It is also referred as *row*, *translation unit* in TMX (the `tu` element), *alignment* in Euramis, etc. There are two field types: segments and non-segments. For example, above the columns *English* and *Spanish* are segment types; the column *Number* is non-segment type, a unique key (MATID in Euramis). It might contain additional non-segment fields such as *domain* (e.g., law), link to the *source file*, etc.

One can create separated linguistic tables for specific non-segment fields and eliminate the corresponding fields; e.g., one might create a table only for the domain law and hence it would not need to contain the classification field indicating the domain.

The main types of linguistic tables are:

- **Translation memory data**: a small linguistic table.
- **Linguistic database**: a large linguistic table.

### 10.6.2 Translation memory data

It is a **small** linguistic table that might contain at most few thousands records. Often, the data is related to one or a small collection of related files. The processing is usually interactive; e.g., with a CAT tool. Translation memory data might be created in batch from a linguistic database; usually, the exchange format is TMX, though one might use any format that represents a linguistic table, not even necessarily in XML (e.g., comma separated values). Also, a translation memory data could be constructed while translating.

Translation memory data beyond a certain size (i.e., when it starts to become a linguistic database) becomes inefficient, particularly with poor or no indexes. It is far better to maintain a linguistic database and extract a small translation memory data specific to one document. Also, one has to be careful with the implementation (see *TMX considered harmful*).

*Translation memory* (TM) refers to the technique (i.e., literal strings) as opposed to traditional rule-based *machine translation* (MT); though TM is closely related to *statistical machine translation* [SMT]. The term TM might be ambiguous and one might have to be more specific: *translation memory* **tools** are programs for processing *translation memory* **data**.

### 10.6.3 Linguistic database

It is a **large** linguistic table that might contain millions of records; e.g., Euramis. Often, the processing on linguistic databases is in batch; also, for extraction (exact or fuzzy) it requires indexes.

### 10.6.4 Linguistic data harvesting

Abbreviated to *harvesting*. It is data extracting (segments and other data) from parallel texts (e.g., parallel corpora, completed or ongoing translations, etc) into linguistic tables. There are two types:

- **Programmatic harvesting**: data extracting with a **batch** program **without** human intervention. Usually it would go directly into a linguistic database.

- **Manual harvesting**: data extracting with an **interactive** program **with** human intervention, usually with processing the text for other purposes; e.g., translation. Usually it would go into a translation memory data and later into a linguistic database.

The additional data might be: a pointer to the file(s) from where the segment was extracted from, if possible to the exact place in the text; domain (e.g., law), etc. These two harvesting types mirror segmentation (also programmatic and manual) and share similar characteristics.

### 10.6.5 Record value

Ideally, this metric should reflect the **future** usefulness. If a person knows this value, it can be set manually; e.g., for a new term that will be used in the future such as the term *blog* when it was coined. More realistically, the value is calculated according to usage. The problem is that *reading* a record is not the same as *using* a record; the metric of reading a record is better than nothing, though it is far from reflecting the real usefulness of the record. In particular, a low value means low usefulness, but a high value does not necessarily mean a high usefulness; in other words, the figure detects low usefulness but not necessarily high usefulness. Analysing the translated documents can help to refine the record value; i.e., one can determine which records were really used, particularly if the format keeps the record identifier.

**10.6.6 One to many records**
A typical case is due to inflexion. Several records might have the same segment content in a language (e.g., English) and different segment contents in another language (e.g., Spanish); the word roots should be the same. The more languages, the trickier it is; e.g., the records 2 and 3 in the table above due to grammatical gender.

An extraction from a linguistic database to create a translation memory might result in several records for one segment; in the case of CAT, a human translator would choose the correct segment. In the case of generation of parallel texts one has to be particularly careful.

## 10.7. Combination of textual and tabular data

**10.7.1 Parallel texts corpora**
Abbreviated to *parallel corpora*. It is a large set of parallel texts files that are not necessary related. Parallel corpora must be properly structured, so it can be processed programmatically and be used as the basic data for creating tabular data. From the raw parallel corpora one could create intermediate parallel corpora; e.g., from a parallel corpora with mostly RTF files it can be created an intermediate parallel corpora with files mostly in XML, so to facilitate further processing. One must preserve the original format as new algorithms might extract more information from the original documents.

Parallel corpora are the **basic ingredient** for techniques such as translation memories and the very promising *statistical machine translation* [SMT]. Daydreaming: one computer disk with a directory structure containing all the translations made by of all the European institutions/bodies: over 56 years worth of translation at present in 23 languages; millions of pages. Indeed, whoever controls such massive repository dominates the field; in comparison the programs are secondary.

**10.7.2 Record identification in text**
In continuous narrative text, segments should be identified if they come from linguistic databases or similar resources. The identification is the linguistic database name and the record number; the segment (the field) as per the *processing language*. This might be done with a *Uniform Resource Identifier* (URI); e.g., `http://example.com/r1`, where `http://example.com` is the database name and `r1` the record number, though the database name could be in a header and applied to the whole file (similar to the `base` element in HTML) and only the record number (`r1`) would be with each segment.

# 11. Electronic dossier

All the elements on a paper-based production of parallel texts constitute a dossier containing administrative data, source text, translations, background documentation, etc. In the electronic world, one also needs an *electronic dossier* with all these components, if one seeks interoperability in the whole ATP-chain, particularly in the case of *loose phase binding*.

Xdossier [XDOSSIER] is a type of electronic dossier. It is a *data object* composed of a directory structure with files often in several formats. It is designed for browsing with web browsers. An Xdossier can be one file (*container format*) or a data object in a server that has the same outward look as the file.

The data object can be packaged as one single file with zip; similar to ODF and *Microsoft Open Office XML* [OOXML]. The main reason for zipping is packaging a structure containing several files into one file. Xdossier can have three forms, according to the content (file or URI):

- **Self-contained**: all components are in the file.
- **Mix**: some components are in the file and others are external (URIs point to these components).
- **External only**: the file contains only external components (URIs point to these components).

The rationale is that one could process (e.g., create or view) an Xdossier with existing generic tools such as an editor (e.g., vi, Notepad), zipping programs or browsers (e.g., Firefox, Internet Explorer). However, it is better to use specific programs such as a computer-aided translation (CAT) tool.

It should be avoided overloading a format; e.g., using namespace in XML to embed additional terminology. It is generally better to keep the overlay data in separate files. The whole could be packaged into an electronic dossier.

# 12. Multilingual Electronic Dossier (MED)

The MED is an Xdossier designed for the whole ATP-chain. The main MED sections are:

- **Header**: data concerning the whole dossier; e.g. administrative data, statistics, etc.
- **Parallel**: parallel texts.
- **Artefacts**: auxiliary items; e.g., translation memories, background documentation.

It is recommended to view an example of a MED to get a better feeling. The MED is one type of glue for loose phase binding; indeed, the first implemented programs should just use MED as an interface. MED is open and non-proprietary. Other formats and protocols with similar functionalities could be specified.

At the end of the ATP-chain, a MED might have data for the whole ATP-chain or just a subset; e.g., the publisher might not be interested in the gory detail of translation.

# 13. Miscellaneous linguistic aspects

## 13.1. Internationalization and localization (I18N, L10N)

*Internationalization* (I18N) means creating *culture neutral* systems apt for *Localization* (L10N) into different cultures; for example, Greece, Canada-French or Belgium-French. Language is one of the aspects, but there are others such as date representation [XLIFF].

## 13.2. Text reusability

Abbreviated to *reusability*. It is a measure of how much could be copied from previous texts and/or translations to translate a new text. It is a continuum between 0 and 1. Reusability 0 means that the whole text has to be translated. Reusability 1 means the whole text could be copied; e.g., a topic.

## 13.3. TMX considered harmful

TMX is useful for data exchange among unrelated systems. It has little sense to create a physical TMX file containing segments when a computer-aided translation (CAT) tool can address a server; e.g., when the CAT tool and the server are in the intranet of an organisation.

For a new file that has to be translated, one can submit the file in batch to a linguistic database system and the result should be an URI (not physical files) that points to a list of record identifiers (not the segments). CAT tools can use the URI, with the appropriate parameters if needed (e.g., language and/or format), to request translation memories (e.g., a language pair or a set of 23 languages) that might be even in TMX. This approach would avoid the creation and management of hundreds of thousands of physical TMX files. It can be implemented as a web service. The results could be ready, constructed on the fly or a combination of both; e.g., as the *man* command in Unix.

This technique can also be used in other cases such as putting in common linguistic resources from different organisations.

## 13.4. Characters

Unicode [UNICODE] must be the reference model for characters, although one can have different encodings as long as they are labelled; e.g., it is perfectly fine to use Latin-1 [L1] and using the *Unicode Transformation Forma*t in 8 bits [UTF8] does not need to be compulsory.

## 13.5. Language labels

### 13.5.1 Language(s) labelling

There are two types of language labelling:

- **Language(s) of the file**: the metadata indicating the language(s) in a file. It could be **several** languages. In the Dublin Core [DC] is the label `language`; also called *language of the intended audience* [IBP].

- **Processing language**: the language at any given point in the file. It can be only **one** language. It is particularly important in the case of multilingual files. In XML, normally the attribute `xml:lang`. Also called *text-processing language*.

### 13.5.2 Mislabelling

The usage of inappropriate codes must be avoided; e.g., the code `ml` [ISO-639-1], sometimes used to label multilingual files, is reserved for *Malayalam*, a language spoken by 36 millions people.

### 13.5.3 Missing language codes in ISO-639-1

Hereby, it is proposed to the *ISO 639/Joint Advisory Committee* adding to ISO-639-1 the following missing codes:

- **Multilingual file:** code `mm`.
- **Undetermined:** code `un` (in ISO-639-2 is `und`).
- **No linguistic content:** code `xx` (in ISO-639-2 is `zxx`).

If these codes are not available in ISO-629-1 and they are needed, people would use anything.

# 14. Administrative organisation

There should be a new organisation to lead and coordinate this initiative. But as many as possible of the activities (specifications, meetings, etc) must be done within (or in close collaboration) with other existing organisations. For example, the IETF, W3C, *Organization for the Advancement of Structured Information Standards* [OASIS], *Localization Industry Standards Association* [LISA], etc.

This new organisation could be similar to the *XMPP Standards Foundation* [XMPP]. The approach must be similar to the IETF [TAO]: open to everybody, participation at a personal level, inform the stakeholders (e.g., vendors, specification organisations and users), create the specifications and programs simultaneously ("*running code wins*"), create a web site, wiki, mailing list, organise meetings, etc.

In the case of closed collaboration, it might take different forms: working groups, workshop, etc. For example, for some aspects one might apply to the IETF to create a new working group, probably in the *Applications Area* [APPS]; for other aspect one might apply to the W3C for a workshop [NLPIX2008] in the context of their annual conference [WWW2008].

# 15. Legal and miscellaneous

## 15.1. Disclaimer

This document represents only the views of the author and not necessarily the views of any other parties. In particular, it does not necessarily represent the opinion of the European Commission, his present employer.

## 15.2. Comments

To send comments to the author and follow-ups see:

```
http://dragoman.org/par
```

## 15.3. Acknowledgment

The author particularly thanks for their comments: H. Fontes, J. Goetschalckx, P. Hernunez, J. Raybaut, I. Schult, C. Stark, C. van der Horst, Z. Zakic and M. Zouridakis. Also commented: P. Alvarez Rubio, A. Blatt, L. de Prins, A. de Vuyst, H. Jenne and A. Spoden. Though any shortcomings are the sole responsibility of the author. Similarities with the procedures of the IETF are not coincidental.

Ello va seco y sin llover